\documentclass{article}
\PassOptionsToPackage{numbers,sort&compress}{natbib}
\usepackage[preprint]{template/neurips_2026}
\usepackage[utf8]{inputenc}
\usepackage[T1]{fontenc}
\usepackage{amsmath,amssymb,booktabs,graphicx,microtype,float}
\usepackage[table]{xcolor}
\usepackage{tikz,subcaption,tabularx}
\usetikzlibrary{calc}
\definecolor{sourcegold}{HTML}{B27B17}
\definecolor{estgreen}{HTML}{0E9E86}
\definecolor{modelrose}{HTML}{B45A83}
\definecolor{modelgreen}{HTML}{087F70}
\definecolor{linkblue}{HTML}{245699}
\definecolor{bestfill}{HTML}{CBE7DF}
\definecolor{secondfill}{HTML}{F6E6C6}
\newcommand{\best}[1]{\cellcolor{bestfill}\textbf{#1}}
\newcommand{\second}[1]{\cellcolor{secondfill}#1}
\usepackage[colorlinks=true,allcolors=linkblue]{hyperref}

\usepackage{xcolor}         
\usepackage{graphicx}
\usepackage{soul} 
\usepackage[most]{tcolorbox}
\definecolor{siggold}{HTML}{B8860B}   
\definecolor{b0blue}{HTML}{1F5FBF}    
\definecolor{estgreen}{HTML}{2CA089}  

\definecolor{bgblue}{rgb}{0.85,0.92,1.0}   
\definecolor{bggreen}{rgb}{0.88,1.0,0.88}  
\definecolor{bgpink}{rgb}{1.0,0.85,0.93}   
\definecolor{bgpurple}{rgb}{0.80,0.85,1.0} 

\definecolor{textbrown}{rgb}{0.65,0.16,0.16} 
\definecolor{textblue}{rgb}{0.1,0.3,0.7}     
\definecolor{textpurple}{rgb}{0.5,0.2,0.5}   
\definecolor{textgreen}{rgb}{0.0,0.5,0.2}    
\definecolor{fgclay}{rgb}{0.51,0.25,0.04} 


\newcommand{\Bz}{\ensuremath{\mathbf{Q}^{0}}}  

\newtcolorbox{kscomment}{
    colback=bggreen,
    colframe=bggreen,
    boxrule=0pt,
    arc=0pt,
    left=4pt,
    right=4pt,
    top=4pt,
    bottom=4pt,
    breakable
}

\title{Event Signature Transfer: Model-Agnostic Forecast Scenario Construction from Historical Events}

\author{%
Karthik Sridhar$^{1,\dagger}$ \quad
Aaditya Jain$^{1,\dagger}$ \quad
Murari Mandal$^{1,2}$ \quad
Saurabh Deshpande$^{1,*}$ \\[4pt]
$^{1}$Birla AI Labs \\
$^{2}$KIIT Bhubaneswar
}

\begin{document}
\renewcommand{\thefootnote}{}
\footnotetext{$^\dagger$Equal contribution.}
\footnotetext{*Correspondence: \{karthik.sridhar, saurabh.deshpande-c\}@oab.adityabirla.com}
\renewcommand{\thefootnote}{\arabic{footnote}}
\maketitle
 \begin{abstract}

Forecasters often know an event is imminent but not the shape, size, or timing of its
effect. We introduce \emph{Event Signature Transfer} (EST), a training-free,
model-agnostic operator that turns a completed past event into an explicit forecast
scenario. EST removes a source event's own trend and seasonality, then scales and retimes
the remaining \emph{event signature} onto a native forecast, preserving the forecast's linked
structure and reducing to it exactly at zero strength. Because it reads only output
quantiles, EST applies to any quantile forecaster, with no training, no model internals,
at transfer time. Across twelve real episodes and ten synthetic scenarios on
Chronos-2, TimesFM~2.5 and Toto~2.0, manually configured EST reduces real-episode WQL
by 21.7–90\% in-sample. On Chronos-2, it leads eleven of twelve matched comparisons
against covariate conditioning, activation editing and raw replay. The operator builds a scenario; it does not estimate its likelihood.
\end{abstract}

\section{Forecasting a known event from a previous response}

We address a common situation: an event is known or anticipated before the forecast horizon, but the response it will induce in the target is unknown. This arises routinely in practice, where baseline forecasts are adjusted
for exceptional circumstances \citep{fildes2009adjustments}, including events such as
promotions and strikes whose timing may be known before their impact
\citep{nikolopoulos2010events}. A completed historical analogue can provide a candidate
response \citep{green2007analogies}, but it cannot simply be replayed: its trajectory also contains the source's own level, trend, and seasonality, which differ from the target's.

Given this setting, zero-shot time-series foundation models (TSFMs) provide a natural forecasting backbone because they generalize across diverse domains without task-specific retraining \citep{das2024timesfm,ansari2025chronos2}, and Chronos-2 can additionally condition on
known future covariates \citep{ansari2025chronos2}. Recent multimodal systems instead
learn to translate textual or contextual information into forecast changes, including
ChronoSteer, TESS, and Aurora \citep{wang2025chronosteer,li2026tess,wu2026aurora}. These approaches either learn such mappings from large multimodal corpora linking context
to temporal responses, whose relationships may not transfer across targets, or have shown
weak sensitivity to text semantics under controlled audits \citep{sridhar2026semantics}.
Numerical steering offers another route, where methods such as time2time intervene in model
hidden states \citep{sanyal2025time2time}. This leaves a practical gap for transferring a
completed numerical analogue directly onto an existing forecast, without retraining
or access to model internals.

\textbf{Our contribution.} We introduce \textbf{Event Signature Transfer (EST)}, a
training-free, model-agnostic operator that estimates the source's ordinary pre-event
background, removes it to obtain an \emph{event signature}, and places that signature
on the target's untouched forecast with explicit controls for strength, delay, and
duration. EST acts only on output quantiles, requires no additional forecast call, and
reduces exactly to the native forecast at zero strength.

We evaluate EST on twelve real episodes and ten controlled synthetic scenarios across
Chronos-2, TimesFM~2.5, and Toto~2.0, comparing it with raw replay, known-future
covariate conditioning, and a time2time-inspired activation intervention. EST constructs
a scenario conditional on a chosen analogue and controls; it does not estimate the
scenario's probability or identify the correct analogue prospectively.
\section{Extract the response, then steer the forecast}
\label{sec:method}

\begingroup
\setlength{\abovedisplayskip}{5pt}
\setlength{\belowdisplayskip}{5pt}
\setlength{\abovedisplayshortskip}{3pt}
\setlength{\belowdisplayshortskip}{3pt}

{
EST keeps source extraction and target forecasting separate until the final
transfer. Let $x_t$, $t=-N+1,\ldots,0$, denote the source history and
$x_i$, $i=1,\ldots,E$, its completed event. Let $y_t$ denote the target history
and $j=1,\ldots,H$ its forecast horizon.} {As described in Figure~\ref{fig:framework}},{source and target run on independent clocks, aligned only at the Step 3 transfer.
}


\begin{figure}[t]
\centering
\resizebox{1\textwidth}{!}{\input{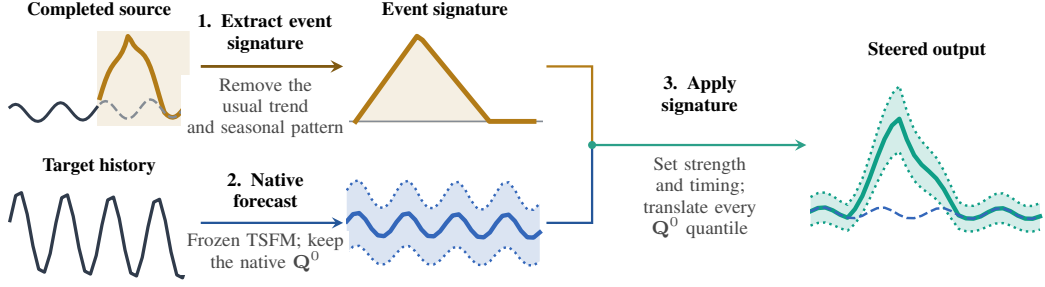}}
\caption{\textbf{Event Signature Transfer Framework.} \textcolor{siggold}{\textbf{Step 1}}: on a
scale where relative changes are comparable, remove the source's usual trend and
seasonal pattern, leaving the event signature. \textcolor{b0blue}{\textbf{Step 2}}: a frozen TSFM
produces the native target forecast \Bz. \textcolor{estgreen}{\textbf{Step 3}}: choose strength and
timing, apply the signature to every quantile of \Bz, and map back to the target's
 scale, giving the steered output.}
\label{fig:framework}
\end{figure}

\paragraph{Step 1: isolate the event from the source background.}{The observed event mixes the response we want to transfer with the source's ordinary
trend and seasonality. We estimate that background from the pre-event history alone, then remove it.} For proportional responses, we work in log space and decompose the source
history into a level, linear trend, and repeating period-$P$ seasonal pattern:
\begin{equation}
 (\hat a,\hat b,\hat s)=\arg\min_{a,b,s}\sum_{t=-N+1}^{0}
 [\log x_t-a-bt-s_{t\bmod P}]^2,
 \qquad
 \sum_{p=0}^{P-1}s_p=0.
 \label{eq:background}
\end{equation}

We extrapolate the fitted background through the event as
$m_i=\hat a+\hat b i+\hat s_{i\bmod P}$ and compare it with the observed source:
\begin{equation}
 r_i=\log x_i-m_i
     =\log\!\left(\frac{x_i}{\exp(m_i)}\right),
 \qquad i=1,\ldots,E.
 \label{eq:residual}
\end{equation}

{The residual $r_i$ is the \textbf{event signature}: how far the source moves above
($r_i{>}0$) or below ($r_i{<}0$) its expected value. We smooth it and keep only the knots
needed within a tolerance, interpolating linearly to give $\hat r(u)$
(Appendix~\ref{app:formal}).}

\paragraph{Step 2: forecast the target independently.}

Using only the original target history, the backbone returns native quantiles
$q^0_{\alpha,j}$ at probability levels $\alpha$, collectively \Bz (baseline target forecast).

\paragraph{Step 3: place the event signature on the target forecast.}

{Three controls set the transfer: strength $w\ge0$, delay $\ell\ge0$ and duration $D>0$.
Defaults $\ell=0$, $D=E$, $w=1$ preserve the source response and its timing; $w<1$
dampens it and $w>1$ amplifies it (Appendix~\ref{app:sensitivity}). For target step $j$,}

\begin{equation}
 p_j=\hat r\!\left(E~\frac{(j-\ell)}{D}\right),
 \qquad
 e_j=wp_j,
 \qquad
 \boxed{q^*_{\alpha,j}=q^0_{\alpha,j}\exp(e_j)}.
 \label{eq:transfer}
\end{equation}

{The term $E(j-\ell)/D$ maps target step $j$ to the source signature; scaling by $w$ and
applying $\exp(e_j)$ to every native quantile injects the event shape into \Bz\ while
retaining the target forecast underneath, giving the steered forecast $\mathbf{Q}^*$
(Appendix~\ref{app:formal}).}

\section{Evidence across events and forecasting backbones}
\label{sec:real}

{\paragraph{Experimental setup.}
We test EST on 12 real episodes across 8 domains (Appendix~\ref{app:selection}) and on
10 synthetic scenarios without real-world confounds (Appendix~\ref{app:synthetic-results}).
For each episode, we manually choose transfer settings from the completed
source response and native target forecast. WQL skill is the percentage reduction in weighted
quantile loss relative to \Bz; higher is better (Appendix~\ref{app:comparators}).}




{\paragraph{EST steers a native forecast.}
Figure~\ref{fig:examples}(b) reads left to right: remove the May 2009 Yampa background, then
place the remaining signature on the untouched May 2023 forecast, which tracks the observed
rise while the target's own structure rides along unchanged. Panel~(a) is the same operation
on an authored dip, where no future truth is defined; Appendix~\ref{app:synthetic-results}
isolates nine more.}


\begin{figure}[t]
\centering
\input{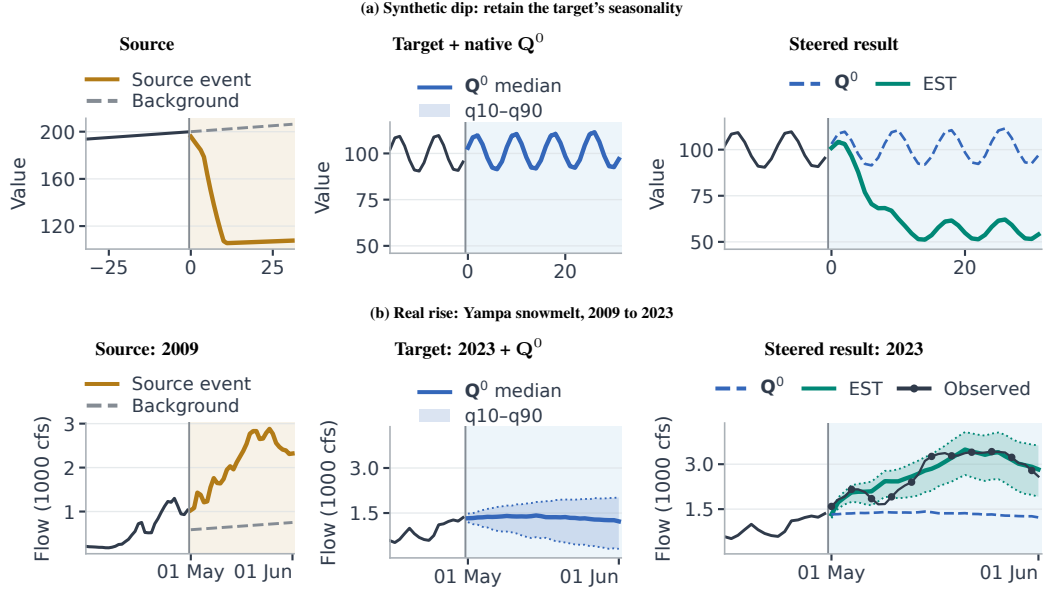}
\caption{\textbf{Source, target, and result.}
(a) An authored dip preserves the target's native rhythm.
(b) The May 2009 Yampa signature steers the May 2023 forecast toward the
observed rise (85.6\% in-sample WQL reduction). Gold denotes the source event and projected background.}
\label{fig:examples}
\end{figure}

{
\paragraph{EST is model-agnostic by construction.}
{EST operates entirely on output quantiles, so it applies to any TSFM, and more broadly any
quantile forecaster, without modification. We run it on Chronos-2, TimesFM~2.5 and Toto~2.0,
which differ in architecture and quantile grid (21, 9, 9 levels)~\citep{ansari2025chronos2,google2025timesfm25,khwaja2026toto2};
the intervention behaves the same on all three (Figure~\ref{fig:models},
Tables~\ref{tab:chronos2}--\ref{tab:toto2}). Never reading internals also avoids the layer
and activation-choice fragility of hidden-state edits.}

\paragraph{EST outperforms existing event-injection methods.}
Table~\ref{tab:real} compares EST on Chronos-2 with three alternatives that also add a
known event to a forecast: raw-source replay, a known-future covariate, and a
time2time-inspired activation edit~\citep{sanyal2025time2time}; each method searches
the controls its own formulation exposes. {Replay directly places a scale-adjusted version of the completed source trajectory on the target, whereas the time2time baseline transfers source activation statistics inside the model; full constructions are given in Appendix~\ref{app:comparators}.}




EST leads eleven of twelve episodes and improves on the native forecast in eleven. The clearest case is refinery, where replaying the raw source worsens the forecast by 140\% while EST improves it by 70\%: replay carries the source's own trend and
seasonality, whereas EST subtracts them and transfers only the residual event. The
two non-wins are honest boundaries, not noise: EST abstains on influenza (too little
history to separate seasonality) and loses narrowly on LA fireworks (the raw holiday
repeats, so removing the source trend adds little).}
\begin{table}[t]
\centering\footnotesize
\caption{\textbf{Chronos-2 in-sample WQL skill (\%) relative to the untouched \Bz\
forecast.} Higher is better; zero matches \Bz. Shading marks
\colorbox{bestfill}{\textbf{best}} and \colorbox{secondfill}{runner-up}.
Constructions and per-backbone results: Appendix~\ref{app:comparators}.
$\dagger$~time2time-inspired adaptation.}
\label{tab:real}
\begin{tabularx}{\linewidth}{*{5}{>{\centering\arraybackslash}X}}
\toprule
Episode & Covariate & t2t-C2$^\dagger$ & Replay & EST\\\midrule
Election attention & +14.1 & 0.0 & \second{+56.4} & \best{+72.7}\\
Gasoline / Harvey & -1.4 & +72.9 & \second{+80.4} & \best{+88.6}\\
Refinery / Harvey & +2.7 & \second{+10.3} & -140.3 & \best{+70.2}\\
Thanksgiving travel & +11.1 & \second{+27.9} & +15.3 & \best{+54.4}\\
Oil / OPEC & -13.4 & \second{+40.4} & +24.1 & \best{+61.5}\\
NYC smoke & +2.4 & +2.1 & \second{+29.7} & \best{+78.7}\\
Michigan claims & -5.5 & 0.0 & \second{+11.8} & \best{+21.7}\\
Influenza-like illness & \second{+36.4} & 0.0 & +28.2 & \best{+56.8}\\
LA fireworks & +1.0 & +6.9 & \best{+65.7} & \second{+61.6}\\
Yampa snowmelt & +27.2 & +6.0 & \second{+74.4} & \best{+85.6}\\
Tax deadline & +51.1 & 0.0 & \second{+63.7} & \best{+79.2}\\
Mumbai monsoon & +9.9 & \second{+14.2} & -479.3 & \best{+45.1}\\
\bottomrule\end{tabularx}

\end{table}


\begin{figure}[H]
\centering
\begin{subfigure}[t]{.35\linewidth}\centering
\includegraphics[width=0.85\linewidth]{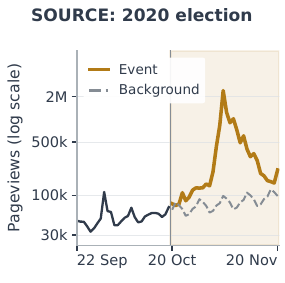}\end{subfigure}\hfill
\begin{subfigure}[t]{.65\linewidth}\centering
\includegraphics[width=0.85\linewidth]{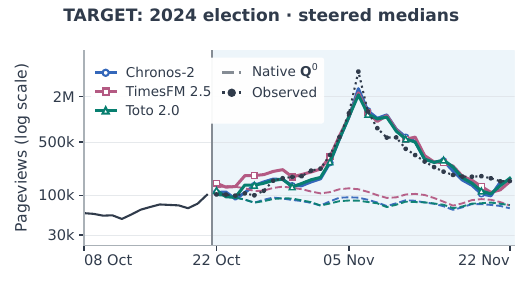}\end{subfigure}
\caption{\textbf{The 2020 election response steers the 2024 election forecast.}
Forecast starts 22 October, before the scheduled 5 November election. Logarithmic axes;
native \Bz\ dashed, steered medians solid. Independent in-sample scenarios reduce WQL by
72.7/64.1/67.9\% for Chronos-2/TimesFM/Toto (Tables~\ref{tab:chronos2}--\ref{tab:toto2}).}
\label{fig:models}
\end{figure}

\section{Conclusion and limitations}
\label{sec:limits}

\paragraph{What EST delivers.}
EST turns a completed analogue into a numerical scenario on top of a frozen forecast,
using output quantiles alone. On Chronos-2 it improves all twelve episodes over \Bz, by
21.7--88.6\% in-sample WQL, and leads eleven of twelve matched comparisons against replay,
covariate conditioning and an activation-patching competitor; the same operator runs
on TimesFM~2.5 and Toto~2.0 without modification, reaching 90.0\% on Yampa snowmelt (Table~\ref{tab:timesfm25}). 

\paragraph{The analyst carries the assumptions.}
EST composes the scenario it is given and does not verify it: the analyst judges that the
event will move the target, picks the analogue and source window, and sets $w$, $\ell$ and
$D$ by inspection. A mismatched analogue is therefore transferred with full confidence,
since the bands carry the native forecast's uncertainty and none about the scenario itself.
Our episodes and controls are chosen with observed outcomes available,
so they show what EST can express, not that the right scenario is identifiable in advance.
Appendix~\ref{app:limitations} treats pretraining exposure, evaluation scope, configuration
choice and analogue mismatch in turn.

\paragraph{Extensions.}
The open problem is choosing and qualifying the scenario before the outcome is known:
LLM-assisted analogue retrieval with $w$ serving as a confidence gate, and an annotated
event-response corpus that supports prospective evaluation
(Appendix~\ref{app:limitations}).
\clearpage
\bibliographystyle{plainnat}
\bibliography{references}
\clearpage
\appendix
\renewcommand{\thetable}{A\arabic{table}}
\renewcommand{\thefigure}{A\arabic{figure}}
\setcounter{table}{0}
\setcounter{figure}{0}
\section{Methodology: from an analogue to a scenario}
\label{app:formal}\label{app:walkthrough}
EST has two inputs: a completed source event with its preceding history, and an
untouched target forecast. The source supplies the response; the target forecast
supplies the baseline. The three steps below make that separation explicit.

\subsection{Extract the source response}
Choose a representation $G_x$ in which the source background is plausibly additive.
On source history alone, fit level, slope and optional seasonality jointly:
\begin{equation}
 (\hat a,\hat b,\hat s)=\arg\min_{a,b,s}\sum_{t=-N+1}^{0}
 [G_x(x_t)-a-bt-s_{t\bmod P}]^2,\qquad \sum_{p=0}^{P-1}s_p=0.
\end{equation}
For $P=0$, omit the seasonal terms. A positive period requires at least $3P$
history samples: three complete cycles. This assumes a stable seasonal pattern
in the chosen representation; it does not discover a period or identify a causal effect.

Extend the background through the completed event and subtract it once:
\begin{equation}
 m_i=\hat a+\gamma\hat b i+\hat s_{i\bmod P},\qquad
 r_i=G_x(x_i)-m_i,\quad i=1,\ldots,E.
\end{equation}
Here $\gamma=1$ continues the source trend and $\gamma=0$ holds its fitted onset
level. Both fit the history slope; the choice concerns its extrapolation.
Only $r_i$ continues to the target. The removed seasonal template has no transfer path.
Errors in the background, however, can remain in the residual.

\paragraph{Keep the useful shape.}
An optional odd-width, order-one Savitzky--Golay smoother gives $\tilde r_i$.
Its interior is a moving average; boundary values use a local fitted line.
Width one leaves the residual unchanged. Add the virtual anchor $\tilde r_0=0$
\emph{after} smoothing. Starting with the endpoints, retain the point with the
largest vertical interpolation error until the piecewise-linear path obeys
\begin{equation}
 \max_{i=0,\ldots,E}|\hat r(i)-\tilde r_i|\leq\tau\max_i|\tilde r_i|.
\end{equation}
The tolerance is relative to the smoothed response amplitude. Explicit source-position
knots may replace this rule; their values still come from the source, but the bound
then need not hold. Smoothing can soften a short peak and is not exact noise removal.

\subsection{Forecast the target, then place the response}
The TSFM receives the original target history once and returns \Bz{} on its native
quantile grid. Request enough forecast samples to cover $H$, then crop to that
same horizon for every comparison. Input and output patch sizes are distinct;
all event controls are measured in time steps.

For source length $E$, target delay $\ell\geq0$ and duration $D>0$, set
\begin{equation}
 p_j=\hat r\!\left(E\frac{j-\ell}{D}\right),\qquad e_j=wp_j,\qquad
 q^*_{\alpha,j}=G_y^{-1}\!\left(G_y(q^0_{\alpha,j})+e_j\right).
 \label{eq:general-transfer}
\end{equation}
The source and target clocks are independent. Set $\hat r(u)=0$ before the source
origin and hold $\hat r(E)$ after its endpoint. A recovered response returns to \Bz;
an unrecovered response retains its final displacement. If $\ell+D>H$, only the
part inside the fixed forecast window is delivered. The API's duration value zero
resolves to $H$; $D=E$ explicitly preserves source timing. Mathematical future
step 1 is plot step 0, and the virtual anchor is plot step $-1$.

An optional cap clips $e_j$ to $[-B,B]$ in target-link units. Every reported scenario
disables it. At $w=0$, EST returns exact copies of \Bz. It never asserts target
samples, rewrites target history, or feeds a steered median back to the TSFM.
\clearpage

\subsection{Choose a meaningful scale and strength}\label{app:sensitivity}
\begin{table}[H]\centering\small
\caption{The link determines what a transferred change means. Source and target may use different links.}\label{tab:links}
\begin{tabularx}{\linewidth}{@{}l l X X@{}}\toprule
Link & $G(v)$ & Target change & Within-step preservation\\\midrule
Log & $\log v$ & Multiply by $\exp(e)$ & Quantile ratios\\
Relative & $v/s$, $s>0$ & Add $se$ & Raw differences\\
Logit & $\log\frac{v-a}{b-v}$ & Multiply odds by $\exp(e)$ & Log-odds differences\\\bottomrule
\end{tabularx}\end{table}
Log requires strictly positive observations and baseline quantiles. Logit requires
values strictly inside declared bounds $(a,b)$. Invalid support and numerical
saturation are rejected. Relative transfer allows signed values and does not
guarantee positive forecasts.

\paragraph{Strength is an assumption about the target.}
The weight $w$ has units of target-link change per source-link change. For equal
log links, $w=1$ transfers the same proportional effect and $w=0.5$ transfers half
the \emph{log response}, not half the raw percentage change. At one step, an
assumed target value $q^\dagger$ implies
\begin{equation}
 w=\frac{G_y(q^\dagger)-G_y(q^0_{\alpha,j})}{p_j},\qquad p_j\ne0.
\end{equation}
For example, an effective source ratio of 1.4 and an assumed target ratio of 1.2
give $w=\log(1.2)/\log(1.4)\simeq0.54$. The effective ratio is $\exp(p_j)$,
after smoothing and retiming, rather than necessarily the raw source/background
ratio. One nonnegative weight constrains the entire path: it cannot reverse the
response or independently match every desired step. At $p_j=0$, only zero
displacement is possible; near zero, the inverse is unstable. With a cap, a
boundary request may admit several weights and a request beyond it is unattainable.
Source amplitude alone does not identify target susceptibility.

For refinery utilisation, a log source maps to a $(0,100)$ logit target.
Weight 8 is a log-odds sensitivity, not an eightfold raw response. At a 90\%
baseline its local proportional sensitivity is $8(1-0.9)=0.8$.
This local derivative is not a finite-change multiplier.

\subsection{What stays intact, and what remains uncertain}
Every quantile receives the same linked displacement. Consequently,
\begin{equation}
 G_y(q^*_{\alpha,j})-G_y(q^*_{\beta,j})
 =G_y(q^0_{\alpha,j})-G_y(q^0_{\beta,j}).
\end{equation}
This preserves native quantile ordering, including any existing crossings.
Removing $e_j$ in link space recovers \Bz. Raw interval widths and raw seasonal
amplitudes need not stay fixed under log or logit transfer. Neither this identity
nor an improved loss establishes event-conditional calibration: the bands omit
uncertainty in the analogue, background, shape, timing and strength.

A wrong period, changing seasonal amplitude or phase, or nonlinear drift can
leave unwanted structure in the signature. Retained controlled source checks
verify cancellation under correct specification and expose mismatches separately.
The ten synthetic scenarios below illustrate mechanisms; they do not supply
forecasting accuracy scores.
\clearpage

\section{Data: events, sources, and response shapes}
\label{app:selection}\label{app:real}
The collection spans attention, energy, travel, air quality, employment, health,
environmental water dynamics (snowmelt and rainfall), and public finance.
Twelve real pairs expose sharp pulses, persistent shifts, troughs with recovery,
and gradual seasonal rises against different backgrounds. Ten authored scenarios
separate these mechanisms. These are individual examples, not a population benchmark.

\subsection{Twelve real episodes}
Tables~\ref{tab:origins} and~\ref{tab:windows} distinguish measurement provenance,
analogue rationale and exact task windows. Public providers supply observations;
cited literature supports response mechanisms. Pairings, windows and scenario
settings are author-defined. Historical event information motivates a scenario
without making its realised magnitude or timing known in advance.
\begin{table}[H]\centering\footnotesize
\caption{Real measurements and the response each analogue is intended to supply. Provider names link to data sources.}\label{tab:origins}
\begin{tabularx}{\linewidth}{@{}p{.32\linewidth}X@{}}
\toprule
Episode and measurement & Analogue and rationale\\\midrule
Election attention\newline\href{https://wikimedia.org/api/rest_v1/}{Wikimedia pageviews} & 2020 election article for 2024. A known election date motivates a concentrated attention pulse. \\
Gasoline / Harvey\newline\href{https://fred.stlouisfed.org/series/GASALLW}{EIA/FRED GASALLW} & Katrina (2005) for Harvey (2017): disrupted supply raises retail gasoline prices~\citep{eia2017harvey,liu2024timemmd}. \\
Refinery / Harvey\newline\href{https://www.eia.gov/dnav/pet/hist/LeafHandler.ashx?n=PET\&s=WPULEUS3\&f=W}{EIA WPULEUS3} & Ike (2008) for Harvey (2017): refinery outage and recovery, opposite in direction to gasoline~\citep{eia2017harvey}. \\
Thanksgiving travel\newline\href{https://www.tsa.gov/travel/passenger-volumes}{TSA checkpoint counts} & Thanksgiving 2023 for 2024: the published holiday calendar motivates a trough and rebound. \\
Oil / OPEC\newline\href{https://fred.stlouisfed.org/series/DCOILWTICO}{EIA/FRED DCOILWTICO} & November 2014 no-cut meeting for December 2015: a sustained price-decline scenario; the policy outcome is uncertain. \\
NYC smoke\newline\href{https://aqs.epa.gov/aqsweb/airdata/download_files.html}{EPA AirData, NYC} & July 2021 transported smoke for June 2023: a pollution pulse, with weather and source-mixture confounds~\citep{magaritz2025smoke}. \\
Michigan claims\newline\href{https://fred.stlouisfed.org/series/MIICLAIMS}{DOL/FRED MIICLAIMS} & 2019 GM strike for 2023 UAW action: anticipated disruption raises claims, but staged and concentrated strikes have different tempos. \\
Influenza-like illness\newline\href{https://cmu-delphi.github.io/delphi-epidata/api/fluview.html}{CDC/Delphi, national wILI} & 2003--04 H3N2 season for 2017--18: a broad illness rise and recovery, motivated by CDC surveillance. No annual background is fitted. \\
LA fireworks\newline\href{https://aqs.epa.gov/aqsweb/airdata/download_files.html}{EPA AirData, Los Angeles} & July 4, 2023 for 2024: a statutory holiday with a sharp particulate pulse~\citep{seidel2015fireworks}. \\
Yampa snowmelt\newline\href{https://waterdata.usgs.gov/monitoring-location/09239500/}{USGS gauge 09239500} & 2009 for 2023: the closest prior April mean flow motivates a snowmelt-rise analogue; snowpack and later weather differ~\citep{stewart2005streamflow}. \\
Tax deadline\newline\href{https://fiscaldata.treasury.gov/datasets/daily-treasury-statement/}{Treasury DTS, non-withheld tax} & 2017 for 2018: align filing deadlines to represent concentrated payments and settlement peaks~\citep{slemrod1997april}. \\
Mumbai monsoon\newline\href{https://power.larc.nasa.gov/}{NASA POWER, PRECTOTCORR} & Mumbai (19.08 N, 72.88 E), 2022 for 2024: a prior near-normal Kerala onset motivates a wet-regime scenario~\citep{wang2002monsoon}. \\
\bottomrule\end{tabularx}\end{table}

\paragraph{Availability matters.}
Published holidays and filing deadlines give a calendar; warnings, advisories
and surveillance give partial event information. An announced OPEC meeting does
not reveal its policy decision. Retrospective AirData, USGS and NASA POWER
snapshots may differ from real-time vintages; POWER precipitation has processing
latency. The Treasury release schedule leaves the first scored day's value
unavailable at the decision. Yampa and Mumbai contexts exclude the previous
annual event. Longer-context comparisons would be different tasks. Data revisions
and possible pretraining exposure remain limitations.
\clearpage

\begin{table}[H]\centering\footnotesize
\caption{Inclusive event and forecast dates. $N_x/N_y$: source/target history lengths; $E/H$: source-event/forecast lengths. Histories end immediately before their respective windows. Oil and tax steps are observed business/trading days; the other clocks are daily or weekly as supplied.}\label{tab:windows}
\begin{tabularx}{\linewidth}{@{}X p{.18\linewidth} p{.18\linewidth}rr@{}}
\toprule
Episode & Source event & Target forecast & $N_x/N_y$ & $E/H$\\\midrule
Election attention & 2020-10-20\newline2020-11-20 & 2024-10-22\newline2024-11-22 & 28/512 & 32/32 \\
Gasoline / Harvey & 2005-08-29\newline2005-09-19 & 2017-08-28\newline2017-09-18 & 156/256 & 4/4 \\
Refinery / Harvey & 2008-09-12\newline2008-11-28 & 2017-08-25\newline2017-11-10 & 156/260 & 12/12 \\
Thanksgiving travel & 2023-11-10\newline2023-12-11 & 2024-11-15\newline2024-12-16 & 42/336 & 32/32 \\
Oil / OPEC & 2014-11-28\newline2015-01-02 & 2015-12-04\newline2016-01-08 & 225/225 & 24/24 \\
NYC smoke & 2021-07-19\newline2021-07-26 & 2023-06-05\newline2023-06-12 & 413/512 & 8/8 \\
Michigan claims & 2019-09-21\newline2019-11-09 & 2023-09-16\newline2023-11-04 & 156/200 & 8/8 \\
Influenza-like illness & 2003-11-16\newline2004-02-29 & 2017-11-26\newline2018-03-11 & 59/260 & 16/16 \\
LA fireworks & 2023-07-01\newline2023-07-08 & 2024-07-01\newline2024-07-08 & 28/336 & 8/8 \\
Yampa snowmelt & 2009-05-01\newline2009-06-01 & 2023-05-01\newline2023-06-01 & 120/200 & 32/32 \\
Tax deadline & 2017-04-03\newline2017-05-04 & 2018-04-02\newline2018-05-03 & 225/225 & 24/24 \\
Mumbai monsoon & 2022-05-28\newline2022-06-28 & 2024-05-28\newline2024-06-28 & 120/240 & 32/32 \\
\bottomrule\end{tabularx}\end{table}

\begin{table}[H]\centering\footnotesize
\caption{Default source extraction. TimesFM ILI instead uses the optional background in Appendix~\ref{app:timesfm-background}. $P$: seasonal period (zero omits it); $k$: smoothing width; $\tau$: shape tolerance; $s_x/s_y$: relative-link scales in original units. No explicit knots or caps are used.}\label{tab:extraction}
\begin{tabularx}{\linewidth}{@{}Xlrlrrr@{}}
\toprule
Episode & Source / target link & $P$ & Trend & $k$ & $\tau$ & $s_x/s_y$\\\midrule
Election attention & log / log & 7 & linear & 1 & 0.04 & -- \\
Gasoline / Harvey & log / log & 52 & linear & 1 & 0.04 & -- \\
Refinery / Harvey & log / logit & 52 & linear & 3 & 0.04 & -- \\
Thanksgiving travel & log / log & 7 & linear & 1 & 0.04 & -- \\
Oil / OPEC & log / log & 0 & linear & 3 & 0.04 & -- \\
NYC smoke & log / log & 0 & linear & 1 & 0.04 & -- \\
Michigan claims & log / log & 52 & linear & 1 & 0.04 & -- \\
Influenza-like illness & log / log & 0 & level & 1 & 0.04 & -- \\
LA fireworks & log / log & 7 & linear & 1 & 0.04 & -- \\
Yampa snowmelt & relative / relative & 0 & linear & 3 & 0.04 & 1000/1000 \\
Tax deadline & relative / relative & 5 & linear & 1 & 0.04 & 1000/1000 \\
Mumbai monsoon & relative / relative & 0 & level & 5 & 0.04 & 10/10 \\
\bottomrule\end{tabularx}\end{table}

\paragraph{Why these representations?}
The source period describes the background removed, not the event's recurrence.
ILI uses all 59 history weeks with $P=0$: Chronos-2 and Toto use the level
background; TimesFM uses the optional source forecast described in
Appendix~\ref{app:timesfm-background}. No annual template is fitted. A 52-week template would require 156 weeks. Yampa's native lower
quantile can be negative, Mumbai's dry baseline is near zero, and tax \Bz{} already
contains part of the payment bump. These motivate relative transfer.
Refinery uses physical $(0,100)$ bounds. The other real targets use log transfer.
\clearpage
\begin{figure}[H]\centering
\includegraphics[width=.32\linewidth]{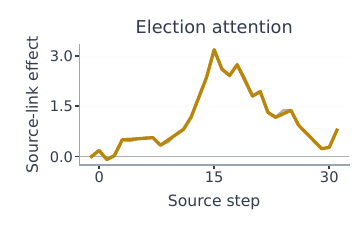}\hfill\includegraphics[width=.32\linewidth]{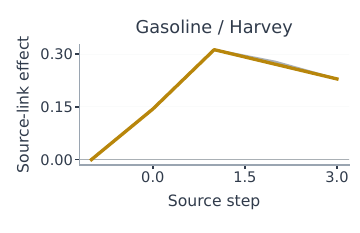}\hfill\includegraphics[width=.32\linewidth]{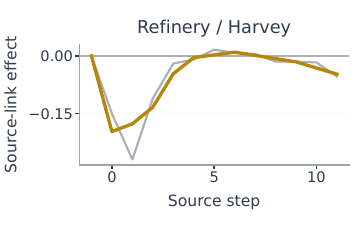}\par\medskip
\includegraphics[width=.32\linewidth]{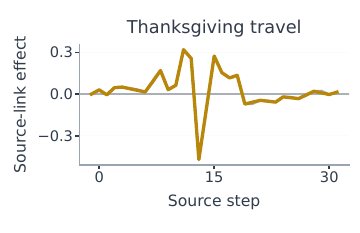}\hfill\includegraphics[width=.32\linewidth]{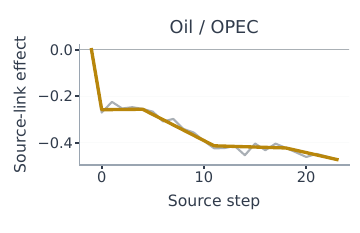}\hfill\includegraphics[width=.32\linewidth]{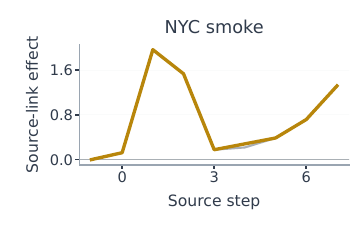}\par\medskip
\includegraphics[width=.32\linewidth]{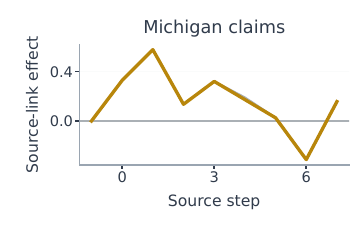}\hfill\includegraphics[width=.32\linewidth]{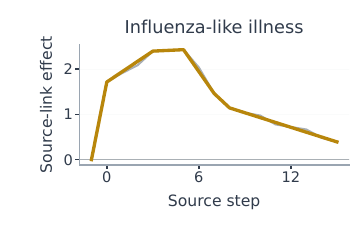}\hfill\includegraphics[width=.32\linewidth]{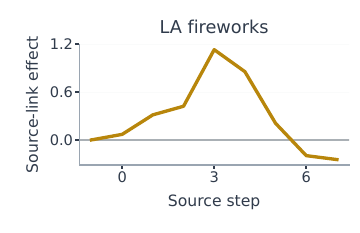}\par\medskip
\includegraphics[width=.32\linewidth]{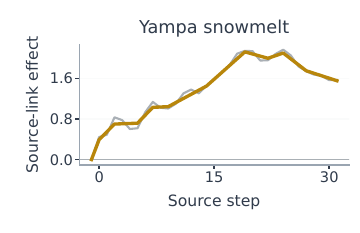}\hfill\includegraphics[width=.32\linewidth]{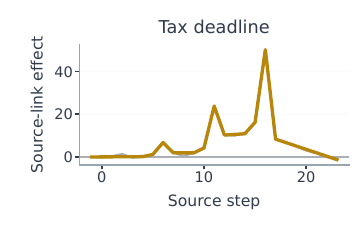}\hfill\includegraphics[width=.32\linewidth]{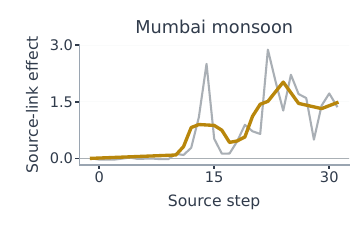}\par\medskip\caption{Twelve source response shapes. Grey is the raw background-adjusted residual; gold is the extracted signature. The virtual zero anchor is at step $-1$. Axes are in each episode\textquotesingle s source-link units and have independent scales. These are source responses, not target truth or causal effects. ILI shows the default signature used by Chronos-2 and Toto; the optional TimesFM background is described in Appendix~\ref{app:timesfm-background}.}\label{fig:real-shapes}\end{figure}

\clearpage

\subsection{Ten synthetic scenarios}\label{app:synth}
Each case has 192 source-history steps, a contiguous 32-step source event,
192 target-history steps and a 32-step forecast horizon. The target future is
intentionally undefined: there is no synthetic truth or accuracy score. All three
TSFMs forecast the same authored histories and receive the same scenario controls.

Let $t=-192,\ldots,-1$ index target history and $u=-192,\ldots,31$ the source.
Ordinary targets are $y_t=100\exp[b_y t+a_y\sin(2\pi t/8+0.3)]$.
The bounded target uses $100\operatorname{sigmoid}[1.4+b_y t+a_y\sin(2\pi t/8+0.3)]$.
Define $z_u=0.001u+a_x\sin(2\pi u/16+0.8)+h_{u+1}\mathbf1\{u\ge0\}$.
Sources are $x_u=200\exp(z_u)$, except the additive-source case, which uses
$200(1+z_u)$ and a relative source scale of 200. There is no stochastic noise.
Extraction receives observations, not the authored background or response.
\begin{table}[H]\centering\footnotesize
\caption{Authored synthetic data and scenario controls. $b_y$ is target slope, $a_y/a_x$ the target/source seasonal amplitudes. $w/\ell/D$ gives strength/delay/duration.}\label{tab:synthetic-defs}
\begin{tabularx}{\linewidth}{@{}Xlrrr@{}}
\toprule
Scenario & Shape & $b_y$ & $a_y/a_x$ & $w/\ell/D$\\\midrule
Downward crash & crash & 0 & 0/0 & 1/0/32 \\
Upward rise & rise & -0.002 & 0/0 & 1.2/0/32 \\
Downward transient shock and recovery & down spike & 0 & 0/0 & 0.8/0/32 \\
Upward transient surge and recovery & up spike & 0 & 0/0 & 1.1/0/32 \\
Crash with the target's weekly rhythm retained & crash & 0.0005 & 0.1/0 & 0.9/0/32 \\
Surge with the target's weekly rhythm retained & up spike & 0 & 0.12/0 & 1/0/32 \\
Rising event; additive source rhythm removed & rise & -0.002 & 0/0.25 & 1.4/0/32 \\
Source period 16 removed; target period 8 retained & down spike & 0 & 0.12/0.2 & 0.8/0/32 \\
Explicitly delayed, compressed surge & up spike & -0.0005 & 0.08/0.1 & 1.3/6/20 \\
Outage and recovery within physical capacity & down spike & 0 & 0.04/0.1 & 2/0/32 \\
\bottomrule\end{tabularx}\end{table}

For normalized event time $v=i/32$, the authored response $h_i$ interpolates
these knots: crash $(0,0),(.15,-.1),(.35,-.65),(1,-.65)$;
rise $h=.5v$; downward transient $(0,0),(.3,-.6),(.65,0),(1,0)$;
upward transient $(0,0),(.35,.7),(.75,0),(1,0)$.
Sampling between knots can soften a peak. Source rhythm has period 16, and target
rhythm has period 8, making their different roles visible.

The first four cases isolate direction and recovery. The next four add target
rhythm or remove source rhythm. The final two change timing and enforce physical
bounds. Complete three-model outputs appear in Section~\ref{app:synthetic-results}.
\clearpage

\section{Results: matched comparisons and complete scenarios}
\label{app:comparators}\label{app:settings}\label{app:figure-refinement}
\subsection{Read each episode against its own baseline}
All real results are in-sample demonstrations with independent settings per
episode and backbone. Analyst judgment motivates the controls; reported settings
were also informed by observed outcomes. The results assess whether EST can express
the event, not whether the settings would have been chosen correctly beforehand.
No skill score is averaged across episodes.

For native quantile set $A$ and truth $y_{1:H}$, define
$\rho_\alpha(u)=u(\alpha-\mathbf1\{u<0\})$. We report
\begin{equation}
 \mathrm{WQL}(q,y)=\frac{2\sum_{\alpha\in A}\sum_{j=1}^H
 \rho_\alpha(y_j-q_{\alpha,j})}{|A|\sum_j|y_j|},\qquad
 S=100\left[1-\frac{\mathrm{WQL}(q^*,y)}{\mathrm{WQL}(q^0,y)}\right].
\end{equation}
Positive skill improves \Bz; zero matches it. Both denominators are nonzero here.
Chronos-2 supplies 21 quantiles; TimesFM~2.5 and Toto~2.0 supply nine deciles.
We compare methods within each backbone and do not rank backbones by these scores.
Median MAE, absolute losses and exact controls are retained with the scenario exports.

\paragraph{The alternatives use the same target window.}
\textbf{Replay} aligns the raw source and scales it by the last history values
$x_0,y_0$. Let $R(u)$ linearly interpolate $(0,1)$ and $(i,x_i/x_0)$ for
$i=1,\ldots,E$, holding endpoint values outside $[0,E]$. Then
\begin{equation}
 \tilde y_j=y_0 R\!\left(E\frac{j-\ell}{D}\right),\qquad
 \tilde q_{\alpha,j}=\tilde y_j\frac{q^0_{\alpha,j}}{q^0_{0.5,j}}.
 \label{eq:replay}
\end{equation}
Replay selects delay and duration without subtracting the source background.
Its evaluation band borrows \Bz's relative quantile structure; it is not learned
replay uncertainty.
\textbf{Covariate conditioning} supplies the completed source as a known-future
related series to Chronos-2, aligned by relative event position with missing-value
padding. Original target history is retained.
\textbf{time2time-C2} transfers source activation means and standard deviations
over context positions through an AdaIN edit. At a chosen layer, let $A_t$ be a
target activation in the selected context span, and let $(\mu_y,\sigma_y)$ and
$(\mu_x,\sigma_x)$ be per-channel population moments over that span and the source
event's context positions, respectively. Our blended edit is
\begin{equation}
 A'_t=(1-\beta)A_t+\beta\left[
 \frac{A_t-\mu_y}{\max(\sigma_y,\epsilon)}\odot
 \max(\sigma_x,\epsilon)+\mu_x\right],\qquad \epsilon=10^{-5}.
 \label{eq:time2time}
\end{equation}
Operations are channelwise. Layer, blend $\beta$ and context span are controls;
other positions, including future readouts, are unedited before the forward pass
resumes. This Chronos-2 adaptation of time2time~\citep{sanyal2025time2time} uses
floored standard deviations and blending, rather than reproducing the published
experiments. A one-patch source has zero empirical style variance.
Covariate and activation results are reported only for the evaluated Chronos-2
interfaces; missing interfaces are not assigned zero skill.
\begin{table}[H]\centering\footnotesize
\caption{Chronos-2: in-sample WQL skill (\%). Best and runner-up are shaded within each row. The native forecast \Bz{} has zero skill. $\dagger$~time2time-inspired adaptation.}\label{tab:chronos2}
\begin{tabularx}{\linewidth}{@{}>{\raggedright\arraybackslash}Xrrrr@{}}
\toprule
Episode & Covariate & t2t-C2$^\dagger$ & Replay & EST\\\midrule
Election attention & +14.1 & 0.0 & \second{+56.4} & \best{+72.7} \\
Gasoline / Harvey & -1.4 & +72.9 & \second{+80.4} & \best{+88.6} \\
Refinery / Harvey & +2.7 & \second{+10.3} & -140.3 & \best{+70.2} \\
Thanksgiving travel & +11.1 & \second{+27.9} & +15.3 & \best{+54.4} \\
Oil / OPEC & -13.4 & \second{+40.4} & +24.1 & \best{+61.5} \\
NYC smoke & +2.4 & +2.1 & \second{+29.7} & \best{+78.7} \\
Michigan claims & -5.5 & 0.0 & \second{+11.8} & \best{+21.7} \\
Influenza-like illness & \second{+36.4} & 0.0 & +28.2 & \best{+56.8} \\
LA fireworks & +1.0 & +6.9 & \best{+65.7} & \second{+61.6} \\
Yampa snowmelt & +27.2 & +6.0 & \second{+74.4} & \best{+85.6} \\
Tax deadline & +51.1 & 0.0 & \second{+63.7} & \best{+79.2} \\
Mumbai monsoon & +9.9 & \second{+14.2} & -479.3 & \best{+45.1} \\
\bottomrule\end{tabularx}\end{table}

\clearpage
\begin{table}[H]\centering\footnotesize
\caption{TimesFM 2.5: in-sample WQL skill (\%). Best and runner-up are shaded within each row. The native forecast \Bz{} has zero skill. Covariate and activation interfaces were not evaluated.}\label{tab:timesfm25}
\begin{tabularx}{\linewidth}{@{}>{\raggedright\arraybackslash}Xrr@{}}
\toprule
Episode & Replay & EST\\\midrule
Election attention & \second{+52.5} & \best{+64.1} \\
Gasoline / Harvey & \second{+79.1} & \best{+84.8} \\
Refinery / Harvey & \second{-103.9} & \best{+67.9} \\
Thanksgiving travel & \second{+19.7} & \best{+46.5} \\
Oil / OPEC & \second{+42.4} & \best{+68.6} \\
NYC smoke & \second{+24.7} & \best{+77.5} \\
Michigan claims & \best{+26.3} & \second{+25.0} \\
Influenza-like illness & \second{-80.8} & \best{+55.0} \\
LA fireworks & \best{+63.9} & \second{+58.4} \\
Yampa snowmelt & \second{+46.0} & \best{+90.0} \\
Tax deadline & \second{+66.1} & \best{+86.1} \\
Mumbai monsoon & \second{-164.7} & \best{+44.1} \\
\bottomrule\end{tabularx}\end{table}

\paragraph{Optional source-forecast background: TimesFM ILI.}\label{app:timesfm-background}
After inspecting the native forecast and observed outcome, we used an optional
background for this episode only: $m_i=\log\tilde x_i^0$, where $\tilde x_i^0$ is
TimesFM's median forecast of the source event from its 59 pre-event weeks alone.
Thus $r_i=\log(x_i/\tilde x_i^0)$ represents departures from the source's anticipated
wave. It is positive during the early rise and negative during the decline.
With $w=0.2$, $\ell=2$ weeks and $D=8$ weeks, WQL falls from 0.09698 to 0.04365
(55.0\% in-sample). This nondefault option requires one additional source forecast;
all target inputs and quantiles remain unchanged. Other scenarios retain the fitted
source background. The historical zero-strength ILI result remains archived.

\begin{table}[H]\centering\footnotesize
\caption{Toto 2.0: in-sample WQL skill (\%). Best and runner-up are shaded within each row. The native forecast \Bz{} has zero skill. Covariate and activation interfaces were not evaluated.}\label{tab:toto2}
\begin{tabularx}{\linewidth}{@{}>{\raggedright\arraybackslash}Xrr@{}}
\toprule
Episode & Replay & EST\\\midrule
Election attention & \second{+52.3} & \best{+67.9} \\
Gasoline / Harvey & \second{+80.6} & \best{+85.6} \\
Refinery / Harvey & \second{-113.5} & \best{+68.9} \\
Thanksgiving travel & \second{+19.5} & \best{+49.1} \\
Oil / OPEC & \second{+45.0} & \best{+65.8} \\
NYC smoke & \second{+27.4} & \best{+74.5} \\
Michigan claims & \second{-1.4} & \best{+25.4} \\
Influenza-like illness & \second{+48.1} & \best{+70.8} \\
LA fireworks & \best{+63.9} & \second{+57.2} \\
Yampa snowmelt & \second{+41.6} & \best{+71.6} \\
Tax deadline & \second{+53.6} & \best{+80.5} \\
Mumbai monsoon & \second{-466.7} & \best{+44.1} \\
\bottomrule\end{tabularx}\end{table}

\clearpage

\subsection{All real forecasts on all three backbones}
Each row uses one episode and its full forecast horizon. Dashed blue is \Bz,
solid green is EST, black is observed truth, and bands show q10--q90.
The short black segment before step 0 is recent target history; complete input
windows are specified in Table~\ref{tab:windows} and retained in the data export.
Plots use the same scenarios as Tables~\ref{tab:chronos2}--\ref{tab:toto2}.
\begin{figure}[H]\centering
\textbf{Election attention}\par
\includegraphics[width=.32\linewidth]{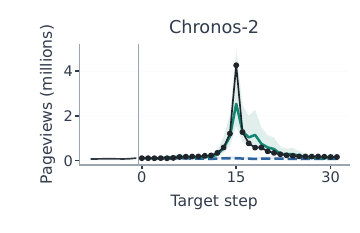}\hfill\includegraphics[width=.32\linewidth]{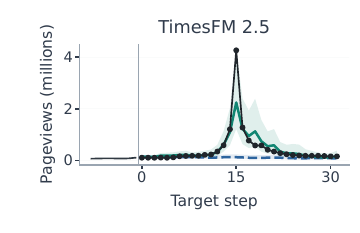}\hfill\includegraphics[width=.32\linewidth]{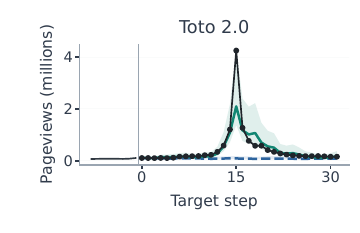}\par\smallskip
\textbf{Gasoline / Harvey}\par
\includegraphics[width=.32\linewidth]{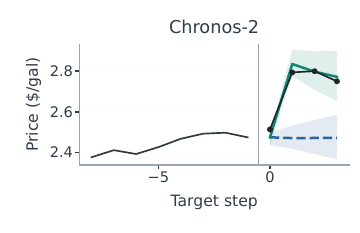}\hfill\includegraphics[width=.32\linewidth]{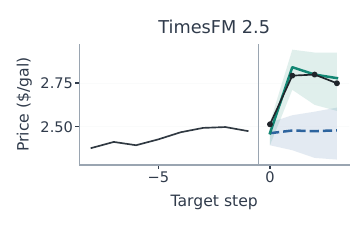}\hfill\includegraphics[width=.32\linewidth]{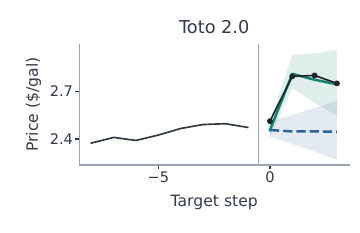}\par\smallskip
\textbf{Refinery / Harvey}\par
\includegraphics[width=.32\linewidth]{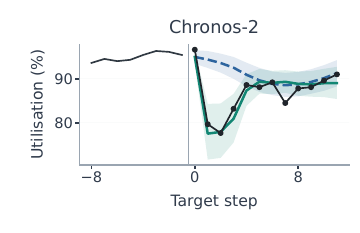}\hfill\includegraphics[width=.32\linewidth]{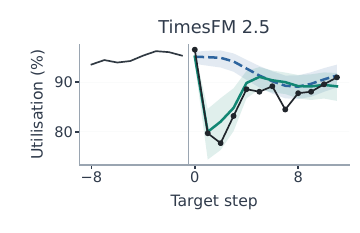}\hfill\includegraphics[width=.32\linewidth]{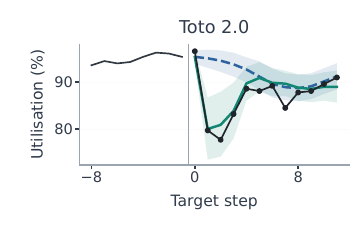}\par\smallskip
\textbf{Thanksgiving travel}\par
\includegraphics[width=.32\linewidth]{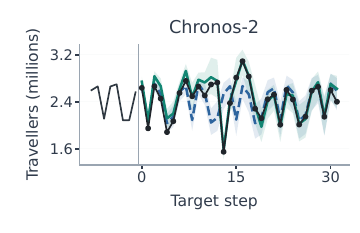}\hfill\includegraphics[width=.32\linewidth]{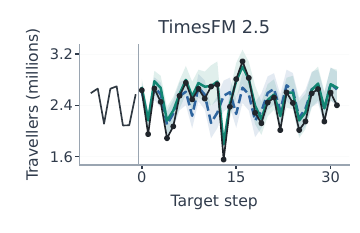}\hfill\includegraphics[width=.32\linewidth]{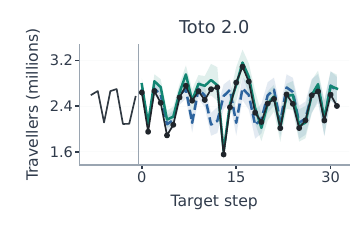}\par\smallskip
\caption{Complete real forecast horizons on all three backbones. Blue dashed: \Bz; green solid: EST; black: observations. Shading: q10--q90. Each model uses its own reported scenario.}\end{figure}\clearpage
\begin{figure}[H]\centering
\textbf{Oil / OPEC}\par
\includegraphics[width=.32\linewidth]{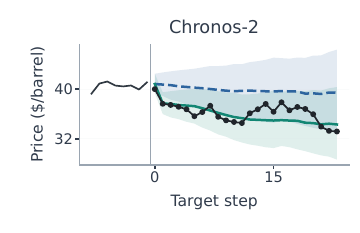}\hfill\includegraphics[width=.32\linewidth]{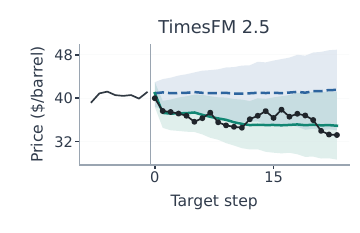}\hfill\includegraphics[width=.32\linewidth]{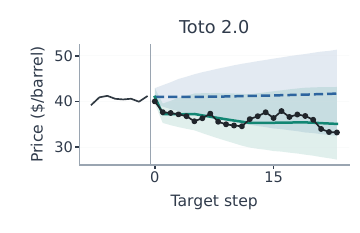}\par\smallskip
\textbf{NYC smoke}\par
\includegraphics[width=.32\linewidth]{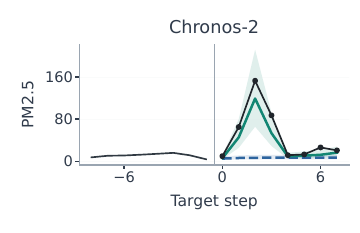}\hfill\includegraphics[width=.32\linewidth]{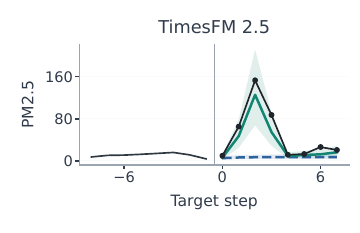}\hfill\includegraphics[width=.32\linewidth]{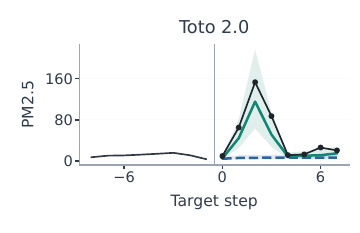}\par\smallskip
\textbf{Michigan claims}\par
\includegraphics[width=.32\linewidth]{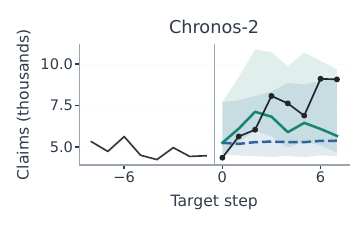}\hfill\includegraphics[width=.32\linewidth]{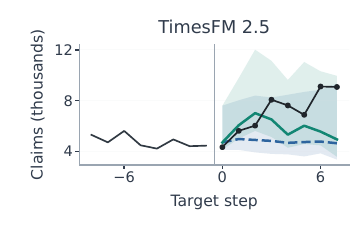}\hfill\includegraphics[width=.32\linewidth]{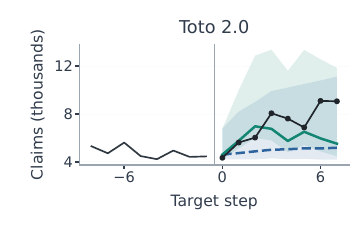}\par\smallskip
\textbf{Influenza-like illness}\par
\includegraphics[width=.32\linewidth]{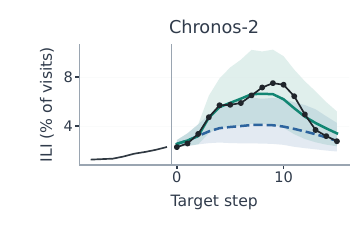}\hfill\includegraphics[width=.32\linewidth]{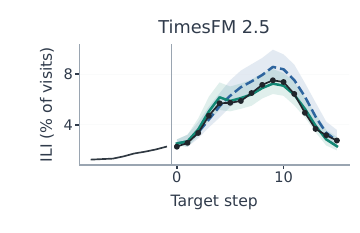}\hfill\includegraphics[width=.32\linewidth]{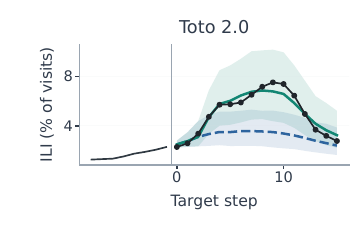}\par\smallskip
\caption{Complete real forecast horizons on all three backbones. Blue dashed: \Bz; green solid: EST; black: observations. Shading: q10--q90. Each model uses its own reported scenario.}\end{figure}\clearpage
\begin{figure}[H]\centering
\textbf{LA fireworks}\par
\includegraphics[width=.32\linewidth]{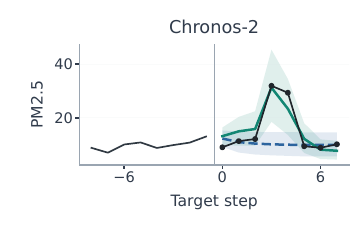}\hfill\includegraphics[width=.32\linewidth]{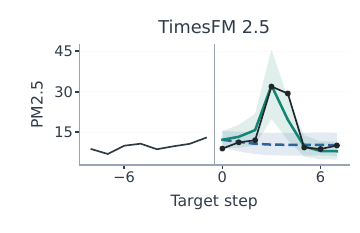}\hfill\includegraphics[width=.32\linewidth]{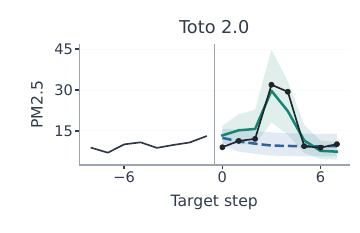}\par\smallskip
\textbf{Yampa snowmelt}\par
\includegraphics[width=.32\linewidth]{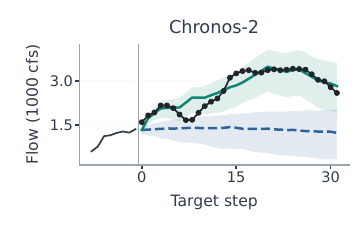}\hfill\includegraphics[width=.32\linewidth]{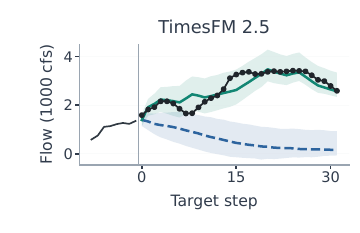}\hfill\includegraphics[width=.32\linewidth]{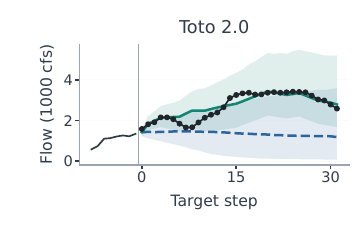}\par\smallskip
\textbf{Tax deadline}\par
\includegraphics[width=.32\linewidth]{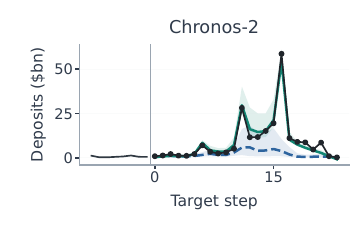}\hfill\includegraphics[width=.32\linewidth]{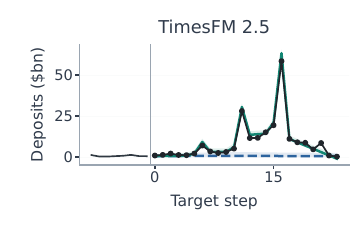}\hfill\includegraphics[width=.32\linewidth]{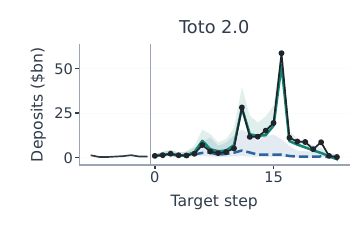}\par\smallskip
\textbf{Mumbai monsoon}\par
\includegraphics[width=.32\linewidth]{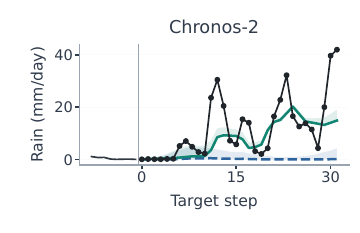}\hfill\includegraphics[width=.32\linewidth]{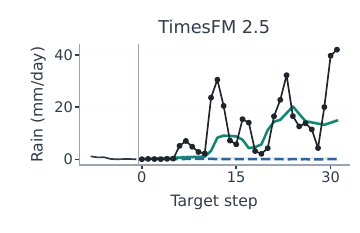}\hfill\includegraphics[width=.32\linewidth]{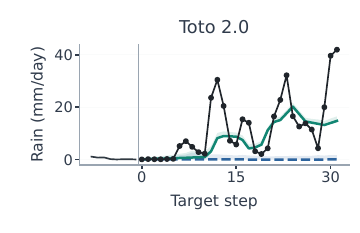}\par\smallskip
\caption{Complete real forecast horizons on all three backbones. Blue dashed: \Bz; green solid: EST; black: observations. Shading: q10--q90. Each model uses its own reported scenario.}\end{figure}\clearpage

\subsection{All synthetic scenarios on all three backbones}
\label{app:synthetic-results}
The source and signature establish the intended response; three model panels
show its composition with different native forecasts. Dashed blue is \Bz{} and solid
green is EST, with q10--q90 bands. There is no black future truth. Source plots use
the last 32 of 192 fitted history steps; forecast plots show the last 16 of
192 target-history steps and all 32 future steps. Display choices do not change
the inputs. Native crossings are retained; counts below each row are \Bz/EST.
\begin{figure}[H]\centering
\textbf{Downward crash and sustained loss}\par
\includegraphics[width=.32\linewidth]{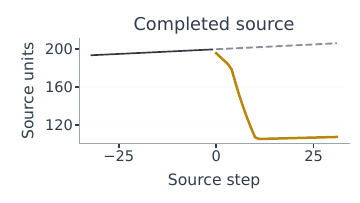}\hfill\includegraphics[width=.32\linewidth]{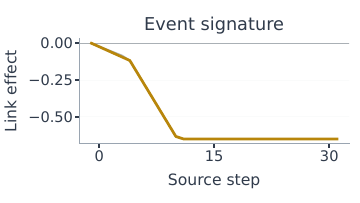}\hfill\includegraphics[width=.32\linewidth]{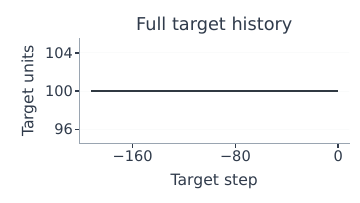}\par
\includegraphics[width=.32\linewidth]{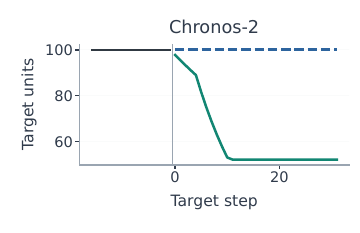}\hfill\includegraphics[width=.32\linewidth]{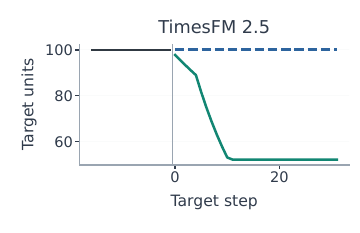}\hfill\includegraphics[width=.32\linewidth]{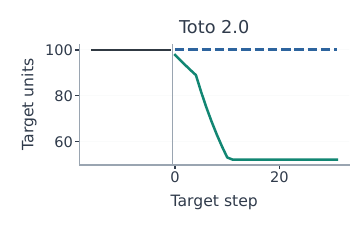}\par{\scriptsize Native/transferred crossings (Chronos-2; TimesFM; Toto): 0/0; 0/0; 0/0.}\par\medskip
\textbf{Upward rise against a declining target}\par
\includegraphics[width=.32\linewidth]{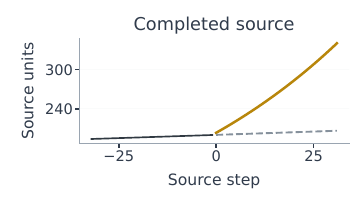}\hfill\includegraphics[width=.32\linewidth]{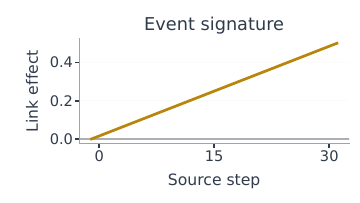}\hfill\includegraphics[width=.32\linewidth]{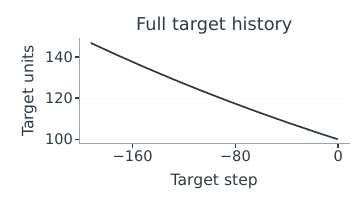}\par
\includegraphics[width=.32\linewidth]{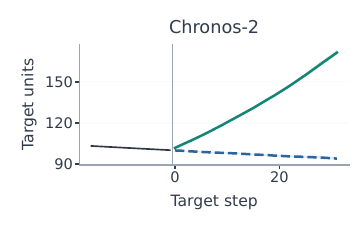}\hfill\includegraphics[width=.32\linewidth]{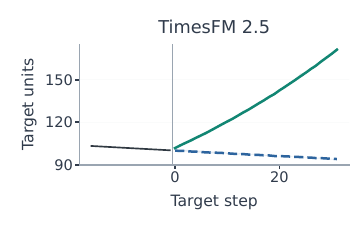}\hfill\includegraphics[width=.32\linewidth]{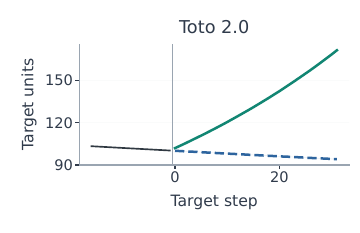}\par{\scriptsize Native/transferred crossings (Chronos-2; TimesFM; Toto): 91/91; 99/99; 0/0.}\par\medskip\caption{Synthetic scenarios: source, signature and full target history above; Chronos-2, TimesFM and Toto forecasts below. No target future truth or accuracy score is defined.}\end{figure}\clearpage
\begin{figure}[H]\centering
\textbf{Downward transient shock and recovery}\par
\includegraphics[width=.32\linewidth]{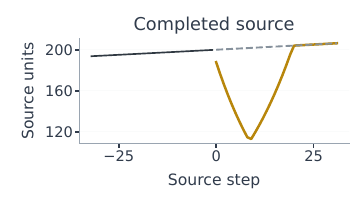}\hfill\includegraphics[width=.32\linewidth]{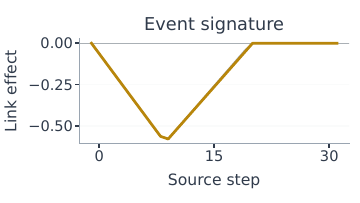}\hfill\includegraphics[width=.32\linewidth]{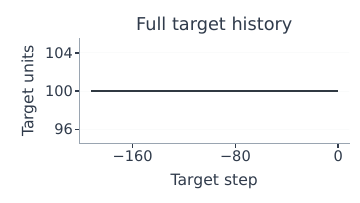}\par
\includegraphics[width=.32\linewidth]{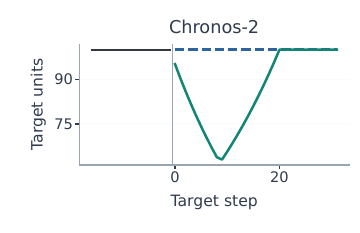}\hfill\includegraphics[width=.32\linewidth]{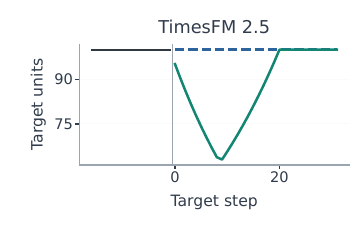}\hfill\includegraphics[width=.32\linewidth]{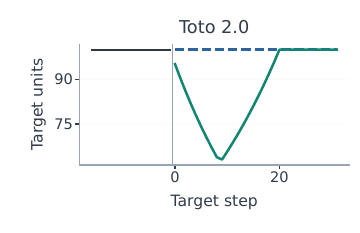}\par{\scriptsize Native/transferred crossings (Chronos-2; TimesFM; Toto): 0/0; 0/0; 0/0.}\par\medskip
\textbf{Upward transient surge and recovery}\par
\includegraphics[width=.32\linewidth]{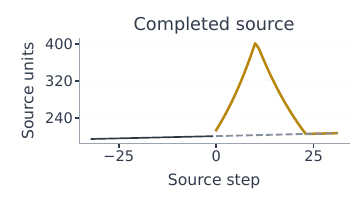}\hfill\includegraphics[width=.32\linewidth]{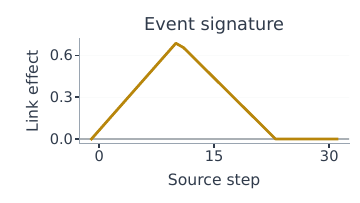}\hfill\includegraphics[width=.32\linewidth]{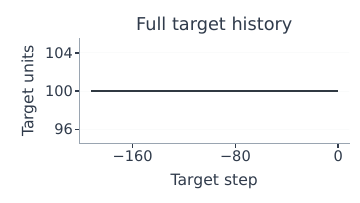}\par
\includegraphics[width=.32\linewidth]{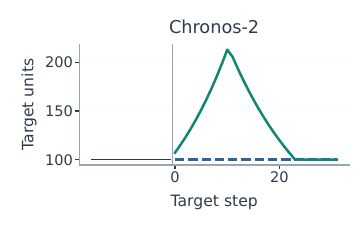}\hfill\includegraphics[width=.32\linewidth]{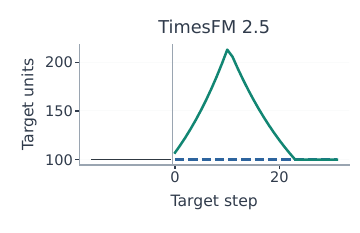}\hfill\includegraphics[width=.32\linewidth]{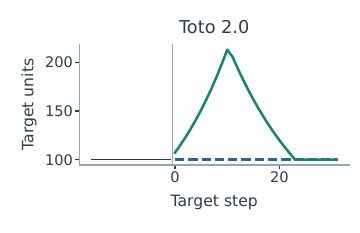}\par{\scriptsize Native/transferred crossings (Chronos-2; TimesFM; Toto): 0/0; 0/0; 0/0.}\par\medskip\caption{Synthetic scenarios: source, signature and full target history above; Chronos-2, TimesFM and Toto forecasts below. No target future truth or accuracy score is defined.}\end{figure}\clearpage
\begin{figure}[H]\centering
\textbf{Crash with the target's weekly rhythm retained}\par
\includegraphics[width=.32\linewidth]{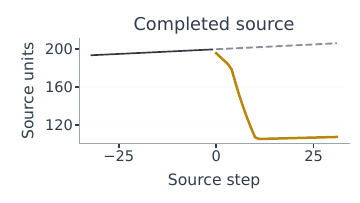}\hfill\includegraphics[width=.32\linewidth]{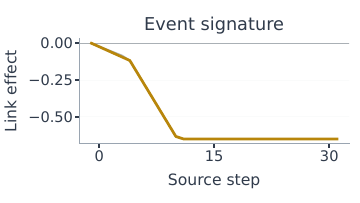}\hfill\includegraphics[width=.32\linewidth]{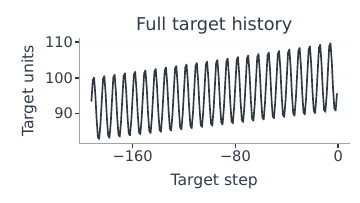}\par
\includegraphics[width=.32\linewidth]{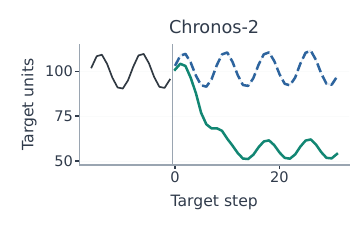}\hfill\includegraphics[width=.32\linewidth]{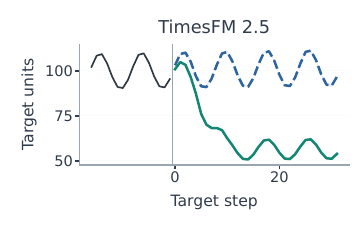}\hfill\includegraphics[width=.32\linewidth]{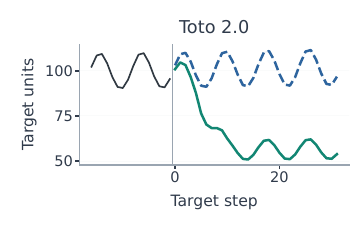}\par{\scriptsize Native/transferred crossings (Chronos-2; TimesFM; Toto): 4/4; 51/51; 0/0.}\par\medskip
\textbf{Surge with the target's weekly rhythm retained}\par
\includegraphics[width=.32\linewidth]{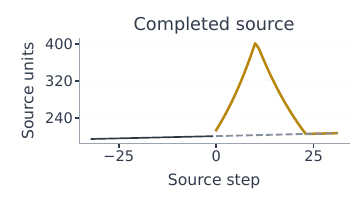}\hfill\includegraphics[width=.32\linewidth]{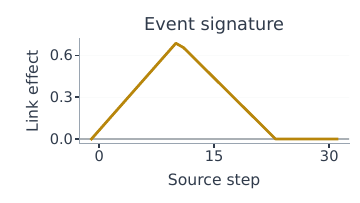}\hfill\includegraphics[width=.32\linewidth]{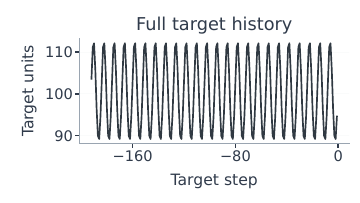}\par
\includegraphics[width=.32\linewidth]{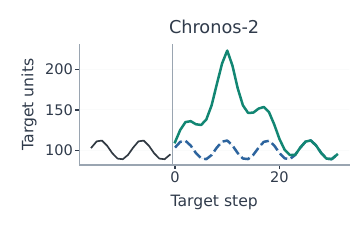}\hfill\includegraphics[width=.32\linewidth]{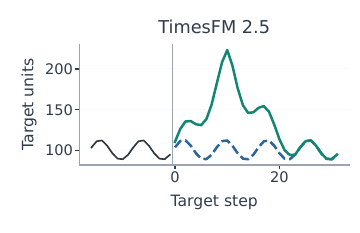}\hfill\includegraphics[width=.32\linewidth]{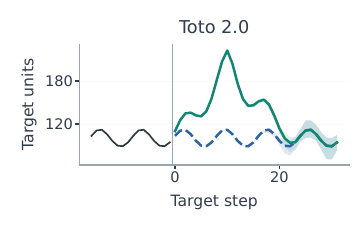}\par{\scriptsize Native/transferred crossings (Chronos-2; TimesFM; Toto): 4/4; 69/69; 0/0.}\par\medskip\caption{Synthetic scenarios: source, signature and full target history above; Chronos-2, TimesFM and Toto forecasts below. No target future truth or accuracy score is defined.}\end{figure}\clearpage
\begin{figure}[H]\centering
\textbf{Rising event; additive source rhythm removed}\par
\includegraphics[width=.32\linewidth]{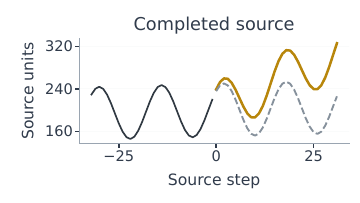}\hfill\includegraphics[width=.32\linewidth]{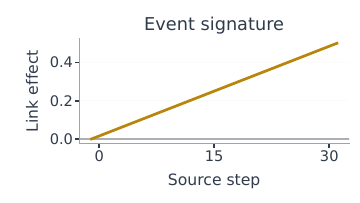}\hfill\includegraphics[width=.32\linewidth]{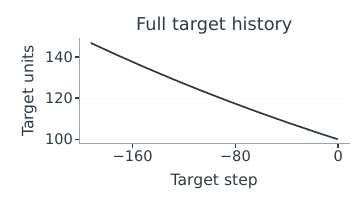}\par
\includegraphics[width=.32\linewidth]{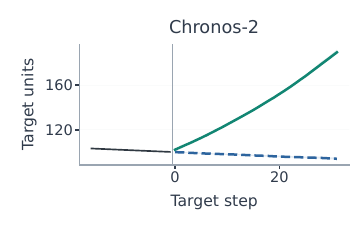}\hfill\includegraphics[width=.32\linewidth]{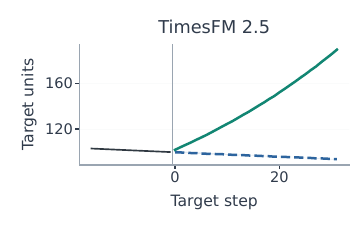}\hfill\includegraphics[width=.32\linewidth]{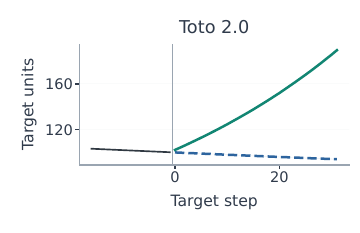}\par{\scriptsize Native/transferred crossings (Chronos-2; TimesFM; Toto): 91/91; 99/99; 0/0.}\par\medskip
\textbf{Source period 16 removed; target period 8 retained}\par
\includegraphics[width=.32\linewidth]{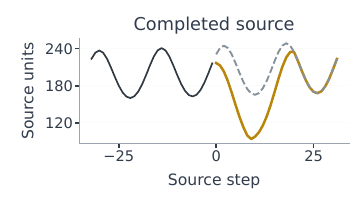}\hfill\includegraphics[width=.32\linewidth]{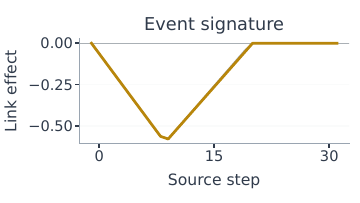}\hfill\includegraphics[width=.32\linewidth]{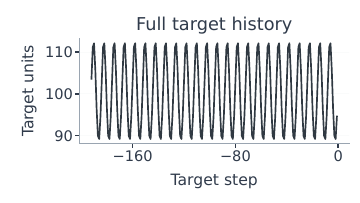}\par
\includegraphics[width=.32\linewidth]{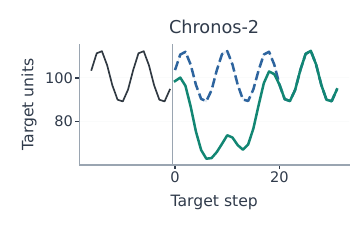}\hfill\includegraphics[width=.32\linewidth]{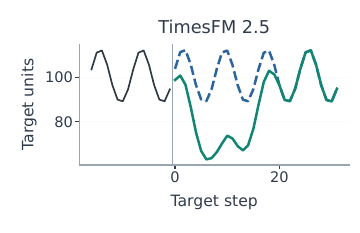}\hfill\includegraphics[width=.32\linewidth]{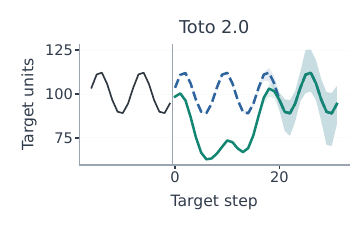}\par{\scriptsize Native/transferred crossings (Chronos-2; TimesFM; Toto): 4/4; 69/69; 0/0.}\par\medskip\caption{Synthetic scenarios: source, signature and full target history above; Chronos-2, TimesFM and Toto forecasts below. No target future truth or accuracy score is defined.}\end{figure}\clearpage
\begin{figure}[H]\centering
\textbf{Explicitly delayed, compressed surge}\par
\includegraphics[width=.32\linewidth]{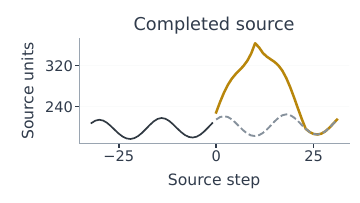}\hfill\includegraphics[width=.32\linewidth]{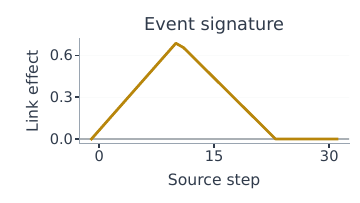}\hfill\includegraphics[width=.32\linewidth]{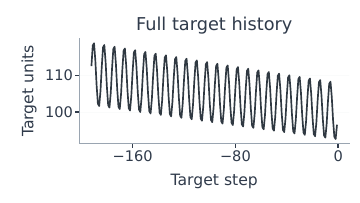}\par
\includegraphics[width=.32\linewidth]{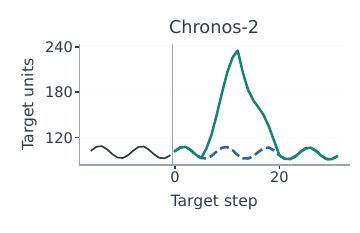}\hfill\includegraphics[width=.32\linewidth]{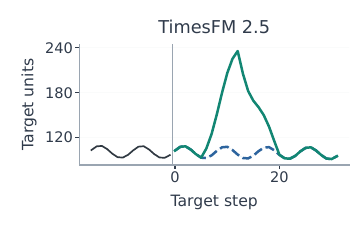}\hfill\includegraphics[width=.32\linewidth]{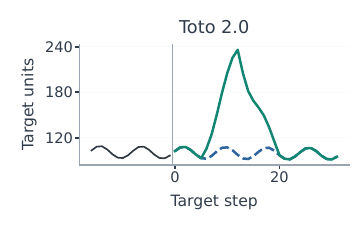}\par{\scriptsize Native/transferred crossings (Chronos-2; TimesFM; Toto): 6/6; 80/80; 0/0.}\par\medskip
\textbf{Outage and recovery within physical capacity}\par
\includegraphics[width=.32\linewidth]{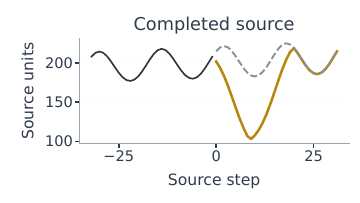}\hfill\includegraphics[width=.32\linewidth]{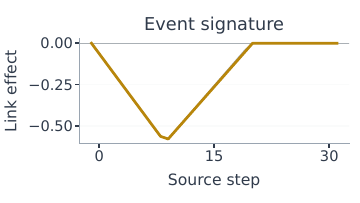}\hfill\includegraphics[width=.32\linewidth]{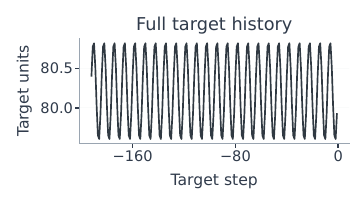}\par
\includegraphics[width=.32\linewidth]{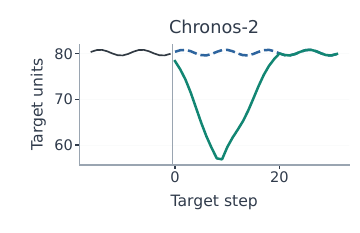}\hfill\includegraphics[width=.32\linewidth]{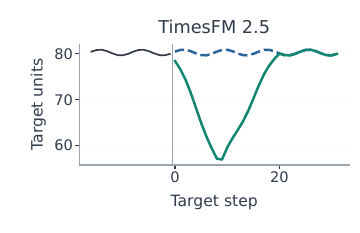}\hfill\includegraphics[width=.32\linewidth]{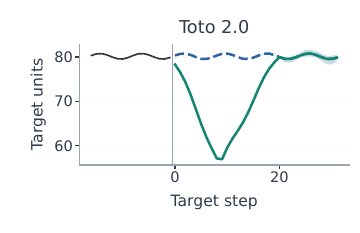}\par{\scriptsize Native/transferred crossings (Chronos-2; TimesFM; Toto): 4/4; 71/71; 0/0.}\par\medskip\caption{Synthetic scenarios: source, signature and full target history above; Chronos-2, TimesFM and Toto forecasts below. No target future truth or accuracy score is defined.}\end{figure}\clearpage

\clearpage
\section{Limitations and extensions}
\label{app:limitations}
EST is a scenario-composition operator, not an estimator of event likelihood. Its output is
conditional on assumptions the analyst supplies, and our evaluation is a set of curated
retrospective demonstrations rather than a benchmark. We set out below what must be assumed,
what our results can and cannot support, and which extensions follow.

\subsection{What the analyst must supply}
\label{app:lim-assumptions}
Four judgments precede any transfer. (i) \emph{Relevance}: that the anticipated event will
move the target series at all. (ii) \emph{Analogue}: a completed episode, in the target's own
history or a related series, whose response mechanism is expected to carry over. (iii)
\emph{Source window}: the span treated as the event, and the pre-event history from which the
background is fitted. (iv) \emph{Transfer controls}: strength $w$, delay $\ell$ and duration
$D$, which state how much of the analogue's response the target is assumed to receive and on
what schedule.

EST makes these judgments explicit without validating their substantive correctness.
The links and background assumptions matter alongside the three controls and source window;
they are recorded in Table~\ref{tab:extraction} and the retained scenario settings, and the operator reduces to the native
forecast exactly at $w=0$. This is the same judgment analysts already exercise when they
adjust a baseline by analogy~\citep{hyndman2021fpp}, moved out of prose and into an auditable,
reproducible form.

\subsection{Pretraining exposure and data leakage}
\label{app:lim-leakage}
Several target windows predate the training cutoffs of one or more backbones, so we cannot
rule out that an episode appeared in pretraining. The matched design helps interpret the comparison but cannot eliminate this uncertainty.

\textbf{Forecast shape does not establish exposure.} Native forecasts differ by episode and backbone: TimesFM already predicts the ILI
wave, while Chronos-2 and Toto substantially underpredict it. Forecast shape alone
cannot establish or exclude pretraining exposure. The optional ILI source background
therefore removes the response anticipated by TimesFM (Appendix~\ref{app:timesfm-background}).

\textbf{The comparison is internal.} Skill is measured against that same backbone's \Bz\ on
the same target window and quantile grid, so any advantage from shared exposure is present in
both forecasts, although its effect on their loss ratio need not cancel. We never rank one backbone against another on these
scores.

\textbf{No target-window observation enters the transfer.} Extraction uses only the source
series and its pre-event history; the target contributes only its own history, through \Bz.
Target truth is used for evaluation and manual scenario selection, never inside the transfer operator. The source is a different year or a
different series from the target in every episode, and the Yampa and Mumbai target contexts
are cut so that the previous annual occurrence lies outside the input window, which prevents
a nearby recurrence from being copied out of context.

Data vintage is an additional caveat: AirData, USGS, NASA POWER
and Treasury snapshots are retrospective and can differ from the vintages available in real
time, and POWER precipitation carries processing latency. These affect the realism of a
prospective deployment, not the validity of the matched comparison.

\subsection{Evaluation scope and selection bias}
\label{app:lim-evaluation}
Our evaluation targets analogue-conditioned scenario construction on fixed event windows.
The published time2time experiments~\citep{sanyal2025time2time} use different backbones
and tasks; our Chronos-2 comparator is an adaptation, not a reproduction.
We assembled twelve episodes across eight domains, covering scheduled events (holidays, filing deadlines),
recurring ones (monsoon onset, snowmelt) and irregular ones (hurricanes, strikes, transported
wildfire smoke), and four response geometries (sharp pulse, persistent shift, trough with
recovery, gradual rise).

Curating episodes and inspecting outcomes limits generalisation. Reporting makes that scope
visible but does not remove selection bias. Every episode is reported individually and no skill score is averaged, so a
favourable case cannot absorb an unfavourable one. Comparator advantages remain in the tables: replay beats EST on LA fireworks across all three backbones and narrowly on Michigan under TimesFM. Every episode runs on three backbones with
different architectures and quantile grids, so a result that depended on one model's
idiosyncrasies would be visible. These checks expose differences within the selected set; they do not establish how
representative that set is.

These results establish that EST can express a range of real event responses across
backbones. They do not establish prospective accuracy, and they are not a population estimate
of how often the method helps. The appropriate fix is a shared benchmark of held-out episodes
with configurations registered before the target outcome is revealed.

\subsection{How the transfer configuration is chosen}
\label{app:lim-config}
We select each episode's controls manually. The procedure is: inspect the extracted signature
to see the analogue's shape and duration; inspect \Bz\ to see the level and rhythm it will be
applied to; judge the plausible magnitude of the target's response relative to the source, and
set $w$ accordingly; where the target's tempo plainly differs from the analogue's, set $\ell$
and $D$ to delay, stretch or compress it. The reported controls specify the scenario retained after this inspection.

Two consequences follow, and we state both plainly. The reported settings were informed by the
observed outcome, so the figures and tables are in-sample fit rather than blind prediction.
And the procedure needs domain knowledge: an analyst must know enough about refinery recovery
or monsoon onset to judge whether a source response transfers at full strength. This is the
cost of a method that composes an explicit scenario rather than inferring one, but it is a real
limitation on who can use EST and on how reproducible a given scenario is between analysts.

\subsection{Analogue mismatch, and why \texorpdfstring{$w$}{w} is a confidence gate}
\label{app:lim-mismatch}
The failure mode that matters is a wrong analogue applied confidently. If the source response
does not transfer, EST moves the forecast away from \Bz\ in a direction and on a schedule that
the target never follows, and the loss is worse than leaving the forecast untouched. Michigan
illustrates this risk: the original saved Chronos-2 transfer worsened WQL by 7.8\%,
whereas the later independently chosen controls improve it by 21.7\%. The 2019 GM strike was a concentrated
walkout and the 2023 UAW action was staged, so the two unfold on different tempos, and the
episode yields EST's smallest Chronos-2 improvement of the twelve. Related mismatches include a
background that was not removed cleanly, so source seasonality or drift rides into the
signature, and an event that \Bz\ already partly anticipates, which is then double counted.
Background removal can help: on Chronos-2 Mumbai, replay's skill is
$-479.3\%$ against EST's $+45.1\%$. This contrast does not isolate background removal
from the other scenario choices or guarantee protection against a poor analogue.

The link-space displacement is controlled by $w$; forecast loss need not vary monotonically. The displacement
applied to every quantile is $wp_j$, and at $w=0$ EST returns \Bz\ exactly and cannot do worse
than the native forecast. This makes $w$ a usable confidence gate rather than only a magnitude
control. An analyst with a compelling analogue transfers at full strength; one with a plausible
but uncertain analogue transfers at $w=0.3$ and applies thirty percent of the
analogue's log response; one with no defensible analogue sets $w=0$ and abstains. The gate is currently set by judgment; $w$ is a response scale, not a probability.
Any mapping from retrieval similarity to strength would need prospective validation.

\subsection{What the bands do not represent}
\label{app:lim-bands}
Every quantile receives the same displacement in link space, so the steered interval inherits the
native forecast's uncertainty structure and nothing else. It does not quantify uncertainty in
analogue selection, source window choice, background estimation, event timing or transfer strength.
A narrow steered band reflects inherited baseline dispersion in the chosen link space,
not evidence that the assumed response will occur. Neither the preservation identity in
Appendix~\ref{app:formal} nor an improved WQL establishes event-conditional calibration.

\clearpage
\subsection{Extensions}
\label{app:lim-extensions}
The open problem is not applying a known signature but choosing and qualifying it before the
outcome is known.

\textbf{Retrieval and control selection in the loop.} Large language models, and emerging
time series language models, could take an analyst's description of the anticipated event
together with the target history and a library of candidate sources, then retrieve relevant
analogues, compare their response mechanisms, propose source windows and suggest $w$, $\ell$
and $D$. Candidate rankings should include a response-mechanism rationale and provenance.
Similarity alone does not establish transferable magnitude: proposed controls and abstention
rules would need evaluation on outcomes withheld from the selection process.

\textbf{An annotated event-response corpus.} EST benefits directly from a searchable library of
completed events labelled with timing, context, affected variables, response window and recovery
pattern. Such a resource would make retrieval systematic, allow retrieval and parameter selection
to be evaluated as tasks in their own right, and enable prospective benchmarks in which the
source and controls are fixed without access to target truth.

\textbf{Richer scenarios.} Multiple analogues combined into a scenario ensemble, multivariate
responses across related series, and uncertainty over $w$, $\ell$ and $D$ rather than point
values are possible extensions. Each scenario can use the same operator, but combining
quantiles does not in general yield mixture quantiles; ensemble calibration and cross-series
dependence require additional modelling.
\end{document}